# Dynamic Fault Characteristics Evaluation in Power Grid

Hao Pei[1], Si Lin[1], Chuanfu Li[1], Che Wang[2], Haoming Chen[3], Sizhe Li[3]

1. Xuzhou Power Supply Bureau, Xuzhou, Jiangsu, China, 221000
2. Nanjing Power Supply Bureau, Nanjing, Jiangsu, China, 210000;
3. Nanjing University of Finance & Economics, Nanjing, Jiangsu, China, 210000

**Abstract**

To enhance the intelligence degree in operation and maintenance, a novel method for fault detection in power grids is proposed. The proposed GNN-based approach first identifies fault nodes through a specialized feature extraction method coupled with a knowledge graph. By incorporating temporal data, the method leverages the status of nodes from preceding and subsequent time periods to help current fault detection. To validate the effectiveness of the node features, a correlation analysis of the output features from each node was conducted. The results from experiments show that this method can accurately locate fault nodes in simulation scenarios with a remarkable accuracy. Additionally, the graph neural network based feature modeling allows for a qualitative examination of how faults spread across nodes, which provides valuable insights for analyzing fault nodes.

## 0 Introduction

The ongoing advancement of the power Internet of Things (IoT) has opened new avenues for leveraging power big data [1]. Concurrently, as the construction of smart distribution networks gains momentum and attracts more investment, it has become critically important to assess the quality of these smart distribution network projects [2].

In recent years, data-driven modeling and artificial intelligence algorithms have developed rapidly, and the distribution network evaluation method based on sample learning has also become a research hotspot. It is an effective way to promote the upgrading of dispatching control technology by using artificial intelligence technology to simulate human thinking, learning massive power grid operation data and experience, discovering rules, forming knowledge and guiding power grid operation [3-5]. There are mainly intelligent methods such as Petri net, artificial neural network, genetic algorithm, rough set decision-making, expert system and data mining [6]. Power system needs to extract key features from unstructured fault data by means of data processing. Because of the electrical connection property of power grid structure, power grid structure can be abstracted into graph structure, which can be used to assist judgment. When power grid fault occurs, it can help dispatchers quickly analyze the cause of the accident, comprehensively grasp the key information of fault treatment, and make auxiliary decisions.

In this research, we employ a graph neural network to detect faults in power grid nodes. By statistically analyzing the interconnections between these nodes, we extract vital features that aid in detection. This approach has demonstrated a high degree of accuracy, achieving 99.53% in identifying faults within a simulated IEEE 10-machine 39-node system.

## 1 Preliminaries

### 1.1 Basic structure of IEEE 10 machine 39 node system

The IEEE 10-machine 39-node system is a well-known regional transmission system network in the field of power systems, also known as the New England 39 node system (new England 39 bus system, NE39BS), whose basic structure is shown in Figure 1. This benchmark network is configured in the New England region of the United States and consists of 39 buses, including 10 generator buses and 19 load buses, which are widely used in the fields of small-signal stability study, dynamic stability analysis, power quality analysis and control. This study is based on the 39-node system, and the data are mainly derived from the simulation of the 39-node system.

### 1.2 Basic Concepts of Graph Neural Networks

In recent years, deep learning algorithm has developed rapidly and has been widely practiced and applied in many fields [7,15-16]. Relying on deep learning technology, graph neural network [8-9] extracts complex features such as topological information and node correlation in graph structure, and solves many related problems in computer vision [10], knowledge graph [11-13] and other fields, which has a wide application prospect. Graph is a data structure composed of vertex sets and the relationship between vertices, which can be expressed as $\alpha_{\text{Grpah}} = (V, E)$. $V$ represents a set of vertices; $E$ stands for the set of relations between vertices, which is usually called edge set. Common graph structure data include social networks, chemical molecular structures, etc. Besides, other forms of data can be converted into graph structure data by certain methods according to structural relations, such as images, texts, etc.

The concept of graph neural network was first put forward by Gori et al. [7] in 2005, and then the model was elaborated in more detail by Scarselli et al. [8]. The graph neural network proposed by Gori et al. [7] draws lessons from the research results in the field of neural networks, and can process the graph structure data, extract and mine the deep-seated features of the graph according to the attributes contained in nodes and edges, and use this feature for further calculation.

When using graph neural network to extract features, firstly, the data to be processed is abstractly transformed into data with graph structure, so that its attribute matrix is $Q$, and an adjacency matrix $G$ is constructed according to the connection relationship between nodes. When constructing an adjacency matrix, if it is an undirected graph, the edge attributes do not contain directional information, and the values of $\boldsymbol{G}_{ji}$ and $\boldsymbol{G}_{ji}$ are both 1 when nodes $i$ and $j$ are connected, and 0 when nodes $i$ and $j$ are not connected. If it is a directed graph, the edge attribute contains direction information. When nodes $i$ and $j$ are connected and $i$ point to $j$, the $G_{ji}$ value is 1, the $G_{ji}$ value is 0, and when they are not connected, the $G_{ji}$ and $G_{ji}$ values are both 0.

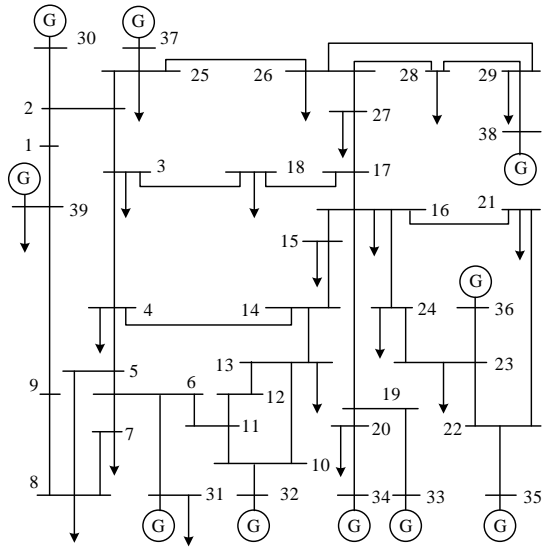

Fig. 1 IEEE10-machine 39-node system

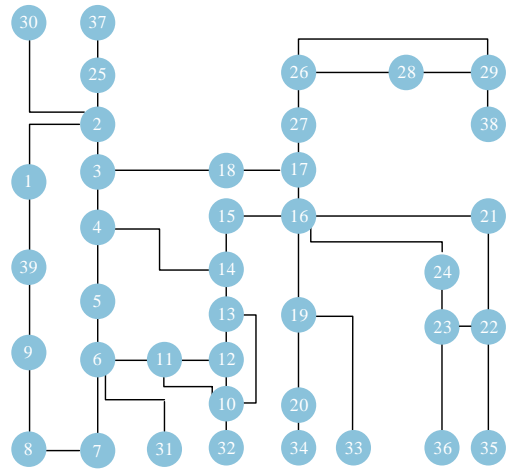

Fig. 2 Topology of the 39-node system of the ieee10 machine.

Initialize trainable weights for each layer of the graph neural network, carry out dimension transformation on the data, and then extract features and learn the correlation between nodes. Wherein the corresponding formula of the graph neural network part can be abstractly expressed as

$$Z_{k+1} = f(Z_k) \quad (1)$$

Where: $Z_k$ and $Z_{k+1}$ are the characteristic matrices of the $k$ and $k+1$ layers; $f$ is the graph neural network layer.

## 2 Grid Feature Extraction

In the related research of power grid fault detection [14], the basic structure of power grid system can be abstracted as graph structure data. By using graph neural network to extract the features of the abstracted graph structure data, the state of each node in power grid system can be detected, and the correlation between nodes can also be explored by using similarity calculation.

### 2.1 Network Structure and Algorithm Flow of Node Fault Analysis

#### 2.1.1 Constructing graph structure data

When analyzing the state of IEEE10-machine 39-node system, a bus can usually be regarded as a node, and the branches between buses constitute the connection between nodes, and then the 39-node power grid system can be defined as Figure G, and the structure is shown in Figure 2.

In IEEE10-machine 39-node system, the node attributes include bus amplitude and phase angle, generator excitation voltage, power angle, active power and reactive power. In order to introduce the characteristics of time dimension, the information of adjacent moments is used to assist the prediction of the state of intermediate moments. The sliding window size of time dimension is set to t moments, and the mean and variance of bus amplitude and phase angle at the current moment and two moments before and after are calculated, which are also used as the attributes of nodes.

Therefore, the graph structure data of the 39-node power grid system can be expressed as $Z \in \mathbf{R}^{39 \times 10}$.

#### 2.1.2 Constructing the neighbor matrix

Regardless of the direction of power flow, the topological graph abstractly generated by IEEE10-machine 39-node system is regarded as an undirected graph. According to the connection relationship between nodes, the adjacency matrix $G \in \mathbf{R}^{39 \times 39}$ is constructed as follows

$$G = \begin{pmatrix} 0 & 1 & 0 & \cdots & 0 & 0 & 1 \\ 1 & 0 & 1 & \cdots & 0 & 0 & 0 \\ 0 & 1 & 0 & \cdots & 0 & 0 & 0 \\ \vdots & \vdots & \vdots & & \vdots & \vdots & \vdots \\ 0 & 0 & 0 & \cdots & 0 & 0 & 0 \\ 0 & 0 & 0 & \cdots & 0 & 0 & 0 \\ 1 & 0 & 0 & \cdots & 0 & 0 & 0 \end{pmatrix}_{39 \times 39} \quad (2)$$

#### 2.1.3 Graph Neural Network Architecture

The architecture of a graph neural network is designed in a U-shaped structure. Initially, the input graph-structured data undergoes an expansion process followed by a reduction stage. The final output features a dimension of 39×1. Each layer within the graph neural network is subjected to a non-linear transformation through an activation function. The operational process of each layer in the graph neural network can be articulated as follows.

$$Z_{k+1} = G \times Z_k \times W_k \quad (3)$$

Where: $Z_k$ and $Z_{k+1}$ are the attribute matrices of the $k$ and $k+1$ layers; $G$ is an adjacency matrix; $W_k$ is the weight matrix of the $k$-th layer.

#### 2.1.4 Loss function

Apply a mask to the output of the network, mask the data of other dimensions except 15 nodes, and get the final predicted value through Sigmoid function, and calculate the loss function by using the final predicted value and label. The loss function adopts binary cross entropy loss ($L_{\text{BCE}}$), and the expression is

$$L_{\text{BCE}} = -\frac{1}{N}\sum_{n=1}^{N} [y_n \log x_n + (1-y_n)\log(1-x_n)] \quad (4)$$

Where: $N$ is the total number of samples; $y_n$ is the probability of true value, which is 0 or 1; $x_n$ is the probability of predicted value.

#### 2.1.5 Node Correlation Analysis in Feature Domain

The analysis of the state of 15 nodes is complemented by examining the relationships between these nodes using features derived from the network. By extracting features from various layers of the network, we perform a cosine similarity analysis. This analysis is then compared to the correlations observed in the original data, serving to validate the utility of graph neural networks in analyzing dynamic characteristics. Specifically, we calculate the cosine similarity between two nodes, represented by their respective feature matrices A and B.

$$S(\boldsymbol{A}, \boldsymbol{B}) = \frac{\boldsymbol{A} \cdot \boldsymbol{B}}{\|\boldsymbol{A}\| \|\boldsymbol{B}\|} \quad (5)$$

## 3 Improvements on GNN

### 3.1 Utilize edge features

The concept of utilizing edge features largely depends on the type of data being used. If the data contains (multi-dimensional) edge features, leveraging these edge features can impact the performance of the model. By using artificial nodes, it's possible to continue using the same model as before. The only change is in the graph itself, where the addition of new nodes makes the graph more complex. Each edge will become a node of its own, connecting to the original deep blue node, rather than being an edge that stores edge features. Indirectly utilize edge features by passing them as node features of synthetic nodes to the model. This can enhance the model's performance if the edge features are relevant to the task. Additionally, consider adding more GNN layers to the model to allow for more neighbor hops.

### 3.2 Self-supervised pretraining

The overall approach is very similar to concepts in the fields of computer vision and natural language processing. Take the language model BERT as an example; this model is trained to predict masked words in sentences (this is self-supervised, as it does not rely on labeled data). We usually don't care much about the specific task of predicting the masked words. However, the generated word embeddings are very useful for many different tasks, as the model really understands the relationships between specific words. Pre-training GNN model node embeddings with a self-supervised task is particularly beneficial in cases with noisy labels, as the self-supervised process provides more "labeled" examples (since no labels are needed for pre-training) and may also be less susceptible to noise interference. The ultimate goal is to classify nodes, and link prediction can be used as a self-supervised pre-training task across the entire graph. Since we are just adding fake edges between existing nodes in the graph and removing real edges, we can do this without relying on any labeled data. The next step is to use the node embeddings from the link prediction GNN model as input for another node classification model.

### 3.3 Separate pretraining and downstream tasks

The GNN model uses self-supervised pre-training to create node embeddings, which are then passed to classic machine learning algorithms or fully connected neural network layers to complete the final downstream tasks. This architecture can be used for many different downstream tasks. The model benefits from combining the ability to access all information contained in the graph with nonlinear manifold learning properties. It inherits some advantages from simpler machine learning algorithms, such as reduced training time and better interpretability.

## 4 Experimental Analysis

### 4.1 Evaluation index

Based on specific parameters like bus amplitude, phase angle, generator excitation voltage, power angle, active power, and reactive power at a certain moment, the state of 15 nodes within the 39-node system of the IEEE 10 machine is assessed. Utilizing these values, the system predicts whether there is a fault present. The effectiveness of this prediction is measured using an evaluation index termed as prediction accuracy.

### 4.2 Experimental Data Set and Experimental Setting

Set the generator parameters of IEEE10-machine 39-node system according to Table 1, and set the stator resistance and reactance to 0, the direct-axis sub-transient reactance and the quadrature-axis sub-transient reactance to 0.2, the additional impedance at the fault of 15 nodes to 0, the load adopts constant power model, and the fault time is set to 0.1 s, and the fault is cut off at 0.70, 0.72, 0.74, 0.76 and 0.78 s respectively to obtain simulation data. The fault data labels generated by cutting faults at 0.70, 0.72, 0.74, 0.76 and 0.78s are set to 1,320 samples in total. Set the non-fault data label of 0.74 s to the final time node, with 927 samples in total, and the data set has 1247 samples in total. The obtained data set is divided into a training set and a test set. The specific division method is as follows: every three times nodes are a group, the first two are set as a training set, and the last one is set as a test set. There are 832 samples in the training set and 415 samples in the test set, with a ratio of about 2:1. The topological structure of all samples in this data set is the same.

After parameter optimization, the model adopts a five-layer graph neural network layer, and is set as a U-shaped structure. The outputs of each layer are 12,18,12,6 and 1 respectively. The deeper features are learned by the way of dimension upgrading first and then dimension reduction, and the outputs of each layer are nonlinear transformed by ReLU activation function. Set the sliding window size $T$ to 5. The model has been trained for 50 cycles, in which the learning rate is set to 0.0001.

Table 1 Generator Parameter Settings

| Generator properties | 1 | 2 | 3 | 4 | 5 | 6 | 7 | 8 | 9 | 10 |
|---|---|---|---|---|---|---|---|---|---|---|
| Synchronizing Reactance of Straight Axis | 0.100 | 0.295 | 0.249 | 0.262 | 0.670 | 0.254 | 0.295 | 0.290 | 0.211 | 0.200 |
| Cross-axis synchronizing reactance | 0.069 | 0.282 | 0.237 | 0.258 | 0.620 | 0.241 | 0.292 | 0.280 | 0.205 | 0.019 |
| Straight axis transient reactance | 0.031 | 0.069 | 0.053 | 0.436 | 0.132 | 0.050 | 0.049 | 0.057 | 0.057 | 0.006 |
| Cross-axis transient reactance | 0.008 | 0.170 | 0.087 | 0.166 | 0.166 | 0.081 | 0.186 | 0.091 | 0.058 | 0.008 |
| Straight Axis Open Circuit Transient Time Constant | 10.20 | 6.56 | 5.70 | 5.69 | 5.69 | 7.30 | 5.66 | 6.70 | 4.79 | 7.00 |
| Cross Axis Open Circuit Transient Time Constant | 0.01 | 1.50 | 1.50 | 1.50 | 0.44 | 0.40 | 1.50 | 0.41 | 1.96 | 0.70 |
| Inertia Time Constant | 42.0 | 30.3 | 35.8 | 38.6 | 36.0 | 34.8 | 26.4 | 24.3 | 34.5 | 50.0 |

## 4.3 Experimental Results

Using the test set to evaluate the network performance, it can be observed that the accuracy is increasing with the increase of training period. When the training period is equal to 45, the value of the loss function is basically unchanged, the model tends to converge, and the final accuracy rate reaches 99.53%. The prediction performance of the model is good, and the accuracy curve and loss function curve are shown in Figure 4. In addition, by using the fault data of other nodes to train the network, the state of each node of the IEEE10-machine 39-node system at a certain moment can be predicted, and the overall operation of the system can be observed in real time.

## 4.4 Ablation experiment

In order to study whether the constructed graph-convolution network model structure is effective, several network framework variants are implemented here, and the performance of each network framework variant is evaluated on the data set, and the ablation experiment is carried out. Table 2 gives the accuracy results of each network framework variant.

Table 2 Test results of network framework variants

| Model | Number of channels | Accuracy/% |
|---|---|---|
| A | 12-18-12-6-1 | 99.53 |
| B | 6-2-1 | 74.64 |
| C | 32-64-32-6-1 | 99.51 |

In Table 2, A is the built network framework, B is the non-U-shaped network with outputs of 6,2 and 1 at each layer of the graph neural network, and C is the U-shaped network with outputs of 32,64, 32, 6 and 1 at each layer of the graph neural network. Although the result of model C is similar to that of model A, its parameters are larger, the calculation amount is also larger, and the model reasoning time is longer. Therefore, the network structure of model A is selected here.

## 4.5 Node correlation dynamic analysis
### 4.5.1 Characterization of node correlation dynamics

Fig. 5(a) shows the top 10 nodes in sliding window 1-3 (including the time when the fault occurs), which have the greatest correlation (calculated by formula (5)) with the fault node (15 nodes), in which the depth of color is positively correlated with the correlation degree, and the deeper the color, the higher the correlation degree (0-1) with the fault node, which is omitted in the figure for brevity. It can be observed that the influence caused by the failure of 15 nodes will have a greater impact on its surrounding nodes in time and space, and the affected areas will also shift in space with time. The graph neural network mainly consists of 4 layers, and the output features of 2-4 layers of the network are extracted for cosine similarity analysis. Figure 5(b) shows the process of feature extraction of the network, and the shallow network features gather in the regional nodes on the right side of the fault node, and with the deepening of the network, the network features gradually gather to the lower left side of the fault node. Through the correlation analysis of 39 power grid nodes in time-space domain and characteristic domain respectively, the propagation of node fault disturbance can be observed qualitatively, which plays an important role in assisting fault node location.

By using the disturbance propagation law of node faults, we can make a preliminary screening when faults occur, narrow the scope of fault node investigation, and improve the efficiency of fault detection and location. Subsequent research can also use the disturbance propagation law to give different weights to node features before they enter the graph neural network to improve the detection accuracy of the network.

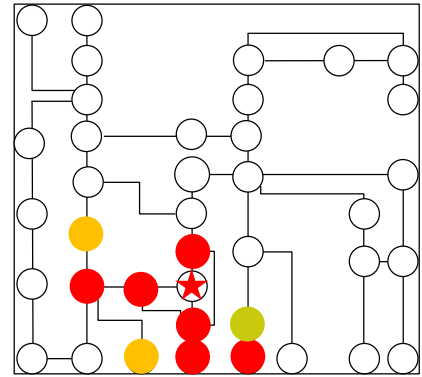

Moment 1

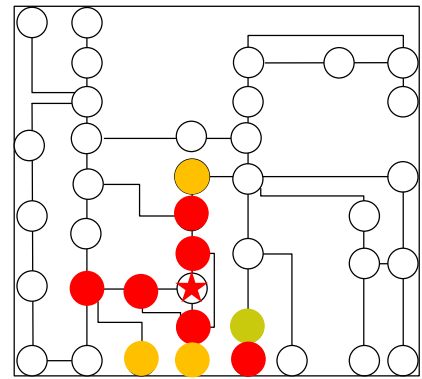

Moment 2

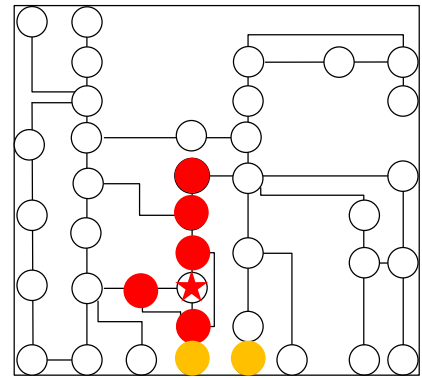

Moment 3

(a) Data correlation in sliding window.

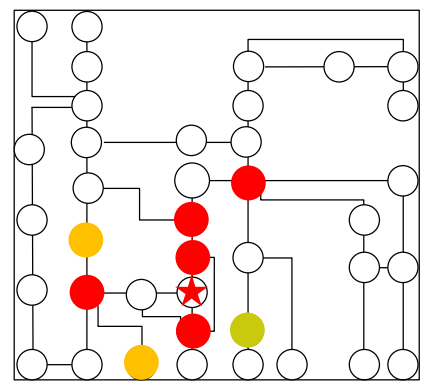

Layer 2 Output

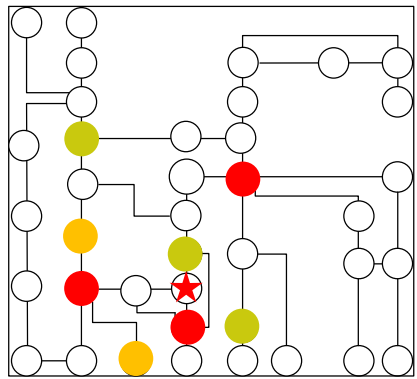

Layer 3 output

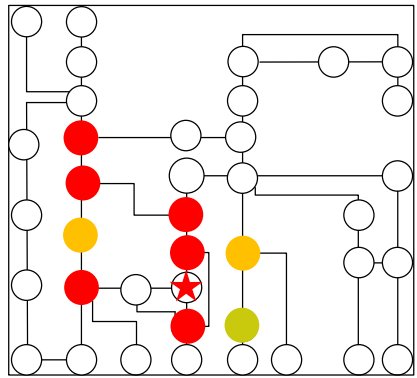

Layer 4 Output

(b) Correlation of features in layers 2-4 of the graphical neural network.

Figure 5 Dynamic characterization of nodes

### 4.5.2 Knowledge Graph Representation for Fault Analysis

By storing these ternary arrays, a knowledge map oriented to fault analysis can be constructed. In order to assist the judgment of fault nodes, it is necessary to comprehensively use the information of multiple nodes. According to the experimental results, the characteristic correlation, time correlation and spatial correlation information of nodes can be used for fusion. According to the experimental results of 15 fault nodes, the information of 8 nodes can be selected for comprehensive judgment, which improves the accuracy of auxiliary judgment. Some important nodes, such as 17 nodes, have strong correlation with fault nodes in feature correlation and spatial correlation, and can be given higher weight for fusion. By comprehensively expressing the information of different bus nodes, we can have a more comprehensive understanding of the characteristics of fault nodes and provide more powerful data support for fault analysis.

## 5 Conclusion

A new model based on graph neural networks is developed for feature extraction in grid systems. This model's performance in fault prediction is validated using simulated data from the IEEE 10-machine 39-node system. Additionally, the model's extracted features are utilized to assess the inter-node correlations in the grid, and these are then compared with correlations derived using traditional methods. Future plans include expanding experiments to larger simulation datasets of the IEEE 10 39-node system, leveraging more extensive data for enhanced node state prediction. Concurrently, the model's practical deployment scenarios will be a focus, with ongoing enhancements to its adaptability for real-world data application.